\documentclass{article}

\usepackage{PRIMEarxiv}

\usepackage[utf8]{inputenc} 
\usepackage[T1]{fontenc}    
\usepackage{hyperref}       
\usepackage{url}            
\usepackage{booktabs}       
\usepackage{amsfonts}       
\usepackage{nicefrac}       
\usepackage{microtype}      
\usepackage{lipsum}
\usepackage{fancyhdr}       
\usepackage{graphicx}       
\graphicspath{{media/}}     

\title{Towards glass-box CNNs}

\author{
  Piduguralla Manaswini \\
  Indian Institute of Technology Hyderabad \\
  \texttt{cs20resch11007@iith.ac.in} \\
   \And
  Jignesh S. Bhatt \\
  Indian Institute of Information Technology Vadodara \\
  \texttt{jignesh.bhatt@iiitvadodara.ac.in} \\
}

\begin{document}
\maketitle

\begin{abstract}
With the substantial performance of neural networks in sensitive fields increases the need for interpretable deep learning models. Major challenge is to uncover the multiscale and distributed representation hidden inside the basket mappings of the deep neural networks. Researchers have been trying to comprehend it through visual analysis of features, mathematical structures, or other data-driven approaches. Here, we work on implementation invariances of CNN-based representations and present an analytical binary prototype that provides useful insights for large scale real-life applications.  We begin by unfolding conventional CNN and then repack it with a more transparent representation. Inspired by the attainment of neural networks, we choose to present our findings as a three-layer model. First is a representation layer that encompasses both the class information (group invariant) and symmetric transformations (group equivariant) of input images. Through these transformations, we decrease intra-class distance and increase the inter-class distance. It is then passed through a dimension reduction layer followed by a classifier. The proposed representation is compared with the equivariance of AlexNet (CNN) internal representation for better dissemination of simulation results. We foresee following immediate advantages of this toy version:  i) contributes pre-processing of data to increase the feature or class separability in large scale problems, ii) helps designing neural architecture to improve the classification performance in multi-class problems, and iii) helps building interpretable CNN through scalable functional blocks. 

\end{abstract}

\keywords{Convolution neural network, Interpretable neural network, implementation invariance}

\section{Introduction}
The learning machines have demonstrated mimicking the human brain in performing some intelligent tasks \cite{ieee_cortex}. They primarily learn large-scale functions from data in many real-life applications \cite{nips,cars,medic,remote}. Many areas like medical diagnosis and security need interpretability of the machine learning (ML) models since human lives are at stake \cite{ieee_medical}. A major challenge is to uncover the multiscale and distributed representation nature of the ML models \cite{explain}. Researchers have approached it through the analysis of visual features, say, saliency maps and variants, or identifying mathematical structures, or by data-driven implementation methods, and/or adding users’ experiences. Here, we choose to perform implementation invariances to gain useful insights for the representational interpretability of CNN. 

Broadly, the learning machines can be categorized into two ideologies \cite{vapnik}: i) the estimated function depends on the features of the data, e.g., k-nearest neighbors (kNN), Naive Bayes, support vector machine (SVM), and the regression; and ii) the estimated function consists superposition of linear indicators arranged to resemble the human's brain, say neural networks and decision trees. Among them, CNNs are suitable for spatially structured data like images and audio signals since they possess shift-invariant and time-invariant properties, respectively. The CNNs layers are of three types viz, convolution layers, pooling layers, and fully-connected layers. Classification by CNNs can be represented as a composition of multiple functions estimation problem wherein each layer estimates an integral function (basket mapping), i.e., $f_1(f_2(f_3(f_4(...))))$, where each $f(.)$ is either a linear or non-linear function parameterized by many unknown parameters. A classification function is typically learned through supervised training with a backpropagation algorithm.

It is found that CNN architecture facilitates rich internal representations of images within the network, thereby, improving the generalization \cite{cvpr_20}. Numerous hyperparameters involved in designing a CNN architecture are the main challenge in vision applications. Besides, the number of filters, sizes of filters, type of activation function(s), and size of the input image are all open questions. By and large, CNNs parameters are primarily adjusted based on the experience of the designer, and the final performance is based on trial and error. Hence, it is hard to produce a guarantee that the optimal structure (CNN model) has been archived. Note that, the user/designer has little to no control over the function learned. This is giving rise to unsubstantiated models which when implemented in sensitive fields like security and healthcare may cause serious issues \cite{bengio_ieee,new_paper}. Hence, the need for interpretable CNN models is imminent \cite{ieee_medical, explain}. In this paper, we perform a post-hoc explanation of the CNN model by analyzing its key properties and rebuilding it through interpretable blocks. In this approach, we move from analyzing the inputs and outputs to a holistic approach to understanding the model behavior irrespective of the input. This way, we are building an interpretable CNN that can help in building better neural models in sensitive fields of medicine and security. The insights gained in this endeavor would contribute to better handling large scale problems using CNNs.

\subsection{Context}
Since the introduction of CNNs, initially, the concern was that the CNNs were finding local minima instead of global minima due to backpropagation \cite{deep}. In \cite{land}, the landscape of function is theorized to have multiple saddle points similar to each other. Later research has been carried out on finding ideal hyperparameters for optimizing its structure, developing various versions of architectures, and understanding its mathematical framework. Many significant statistical methods are introduced into CNNs like probability distributions in lieu of finite, fixed valued weights, and incorporating naive Bayes into deep learning \cite{uncertainity} to improve their performance. Residual network architecture (ResNet) \cite{resnet} uses residual blocks to ease the training process. In an attempt to understand CNNs, the inverted CNN approach is investigated in \cite{invert}. In this, from the last layer (the output), the image is reconstructed by traversing the CNN in reverse order. Such images are then studied to appreciate the framework of CNNs. It is concluded that information about invariance in geometric and photometric data, is learned by the CNNs. Intra-class knowledge inside CNNs is studied in \cite{intraclass}. It is found that CNNs capture location variations and style variations in a class for better classification. 

In recent years, invariance \cite{learn,nips}, and equivariance \cite{rot_eq,ECCV1} in CNNs have been the focal point for analytical studies of deep CNNs. Many researchers have noticed that the filters obtained through training a model are powerful equivariant to group transformations like translation, scaling, rotation, and small diffeomorphisms while non-expansive in nature. Invariant scattering convolution networks (ISCN) are introduced in \cite{ST}. These networks employ wavelets of different degrees of rotations in order to obtain an invariant representation of the image to translation and rotation transformations. They produce outputs at the end of each layer, however, note that the filters in \cite{ST} are predefined wavelets and are not learned from the data. 

More recently, this discovery led to architectures being designed with the idea of built-in equivariance to transformations in group equivariant convolutional networks (G-CNNs) \cite{GECN}. It implements rotated filters in the CNN architecture for equivariant representation to rotation transformations. The input images are also rotated inside CNN for better generalization. It is found to exhibit increased expressive capacity when compared to conventional CNNs with the same number of parameters. The translation equivariance present in conventional CNNs is exploited to include rotation and reflection symmetry. In \cite{PH}, group equivariant non-expansive operators (GENEOs) are introduced and act as filters in CNN. The generated GENEOs are sampled based on a topological data analysis tool.

\subsection{Scope}

In this paper, we perform a post-hoc study of the internal representational aspect performed by CNN layers (basket mapping) and implement the learnings to construct an interpretable prototype for the two-class image classification task. Experiments on MNIST, fashion MNIST, CIFAR-10, plant pathology, skin lesions, cats vs dogs, and cancer detection datasets exhibit encouraging results. The lessons learned during this endeavor would help in addressing real-world large scale problems with scarce data using CNN as follows,
\begin{enumerate}
    \item Pre-processing large scale data to improve the features of class separability.
    \item Improve classification accuracy in multi-class problems.
    \item Towards the design of analytical CNN with higher interpretability.
\end{enumerate}

We sincerely hope that a comprehensive understanding of these popular neural models and their drawbacks would further help in implementing deep learning models in sensitive fields like computer vision in security and health care.

\section{Method}
In this section, we first analytically study the properties of each layer in the conventional black-box CNNs and utilize these insights to rebuild the CNN as a more analytical architecture.
\subsection{The problem}
We set-up the problem using following nomenclature \footnote{ Nomenclature: \begin{itemize}  \item $\{\mathbf{x} \in X\}$, the set of all data points
    \item $\{D\}$ set of all labels
    \item $\{\Phi\}$ set of internal representations of the model\end{itemize}}:
\label{sec:setup}

\begin{itemize}
    \item $f(x) : X \rightarrow D $, function that maps data point to its class label
    \item $\phi(x): X \rightarrow \Phi$, internal representation of a neural model
    \item $f'(x): \Phi \rightarrow D$, classifier that maps internal representation to class label 
    
\end{itemize}
\begin{figure}[h]
  \centering
  \includegraphics[scale=0.3]{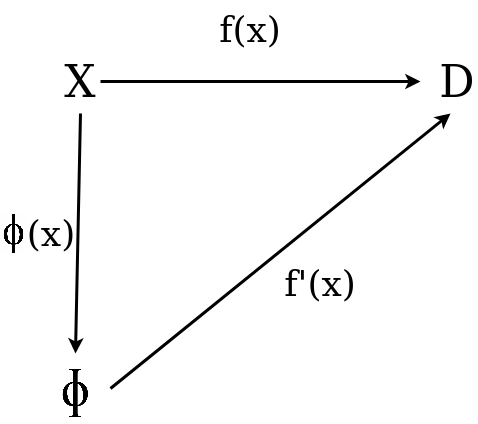}
  \caption{Function map of the problem.}
  \label{fig:space}
\end{figure}
Referring to Fig~\ref{fig:space}, we consider $\mathbf{x}$ as the data with dimension $d$, for example, an image i.e., $\mathbf{x} \in \mathbb{R}^{d \times d}$. A set of all available images is denoted by $\Omega$ which is considered as a subset of $\mathbb{R}^{n \times n}$. The function $f(\mathbf{x})$ maps an $\mathbf{x}$ to its class label $D$ and this is the function one estimate through classification models in deep learning, $f(\mathbf{x}) : \mathbf{x} \rightarrow D$, where $D \in \mathbb{R} $ is set of all class labels. We consider $\Phi = \phi(x)$ as the internal representation of the data in a CNN. Note that such internal representation promotes the generalization ability of a CNN. Consider $g \in G$ is a group of symmetries acting on the function $f$, then $g(x)$ is the transformed image obtained by applying a group transformation $g$. In turn, $f'$ maps the internal representation $\phi(x)$ to the class label, i.e., $f'(x): \Phi(x) \rightarrow D$ (Fig~\ref{fig:space}).

CNNs learn this large scale function $f(x)$, by finding a wide range of representations of the image $\mathbf{x}$ in their internal representation $\Phi$. This $\Phi$ in CNNs is unknown and is a very high dimensional representation. In this paper, we propose a $\Phi$ consisting purely of equivariant and invariant representations followed by dimension reduction. This proposed $\Phi$ is interpretable and is a low dimensional representation. The equivariance ability of the proposed function $\Phi$ is then measured and compared with the well-known AlexNet.

\subsection{Foundation}
\label{sec:insight}
CNNs alternatively apply convolution (linear) and nonlinear operators on the data, layer-by-layer (basket mapping). The linear layer of CNNs contains filters of varying sizes that operate on the output of the previous layer while non-linear layers act as the activation function on the convolved output.

Referring to the problem shown in Fig~\ref{fig:space}, we find a function $\phi$ that separates $f(\mathbf{x})$ such that $f(\mathbf{x})=f'(\phi(\mathbf{x}))$. The function $\phi$ should satisfy the condition if $f(x) \neq f(x^\prime)$ then $\phi(x) \neq \phi(x^\prime)$ where $x,x' \in \mathbf{x}$. There exist two ways to find such a function $\phi$: separation and linearization. In separation, the approximated $f(\mathbf{x})$ using a low dimensional $\phi(\mathbf{x})$ does not decrease the final dimension significantly. It should be noted that in most cases, only one dimension is decreased when separation is implemented. The alternate strategy is to linearize the variations of $f$ with change in variable $\phi(\mathbf{x}) = \{ \phi_k(\mathbf{x}) \}_{k \leq d^\prime}$ where $d^\prime >>> d$, e.g. support vector machine (SVM). It is evident from the success of neural networks that linearization is a better alternative as it leads to better performance. Following this argument, we employ linearization in the proposed interpretable CNN (see Sec~\ref{sec:frame}). The dimension of the data is linearly increased by producing invariant and equivariant representations (see Sec~\ref{sec:equ}) through convolution. 

To linearize the function $f$, we need to find the direction in which the value $f(\mathbf{x})$ does not change. This is achieved by considering level sets, i.e., $\Omega_t = \{\mathbf{x}:f(\mathbf{x})=t\}$ and using groups of symmetries. In this work, we address the two-class image classification problem, hence, we resort to translation, diffeomorphism, and scaling group actions to account for the properties of CNN filters. Since global symmetries are difficult to find, we seek local symmetries $G$ that the function $f$ is locally invariant to :
\begin{equation}
\forall \mathbf{x} \in \Omega,\textnormal{  } \forall g \in G, \textnormal{  } \exists C_x> 0, \forall |g|_G < C_x, f(g(\mathbf{x}))=f(\mathbf{x}). 
\end{equation}

\noindent Here $C_x$ is the extent of transformation that does not impact the recognition of the image. It is the equivariance capability of a representation. Translation group 
\begin{equation}
G \in \mathbb{R} \textnormal{  is   } g(\mathbf{x}(u))=\mathbf{x}(u-g(u)) \textnormal{  with  } g \in \textbf{C}^1(\mathbb{R}^n),
\end{equation}

\noindent where $u$ is the pixel position in the image. Diffeomorphism $diff(.)$ is a stronger constraint where the group transformations are of type diffeomorphism group, i.e.,

\begin{equation}
    G=diff(\mathbb{R}^n), 
\end{equation}

\noindent that transforms $\mathbf{x}(u)$ with differential wrapping of $u \in \mathbb{R}^n$. The properties of group symmetries in the data are combined for estimating the direction of $f(x)$ used while deriving equivariant and invariant representations in this work. Modulus operation on the invariant representation acts as a non-linear component of the representation.
\subsection{A prototype for representational interpretable CNN}
\label{sec:frame}
\begin{figure*}[h]
  \centering
  \includegraphics[scale=0.2]{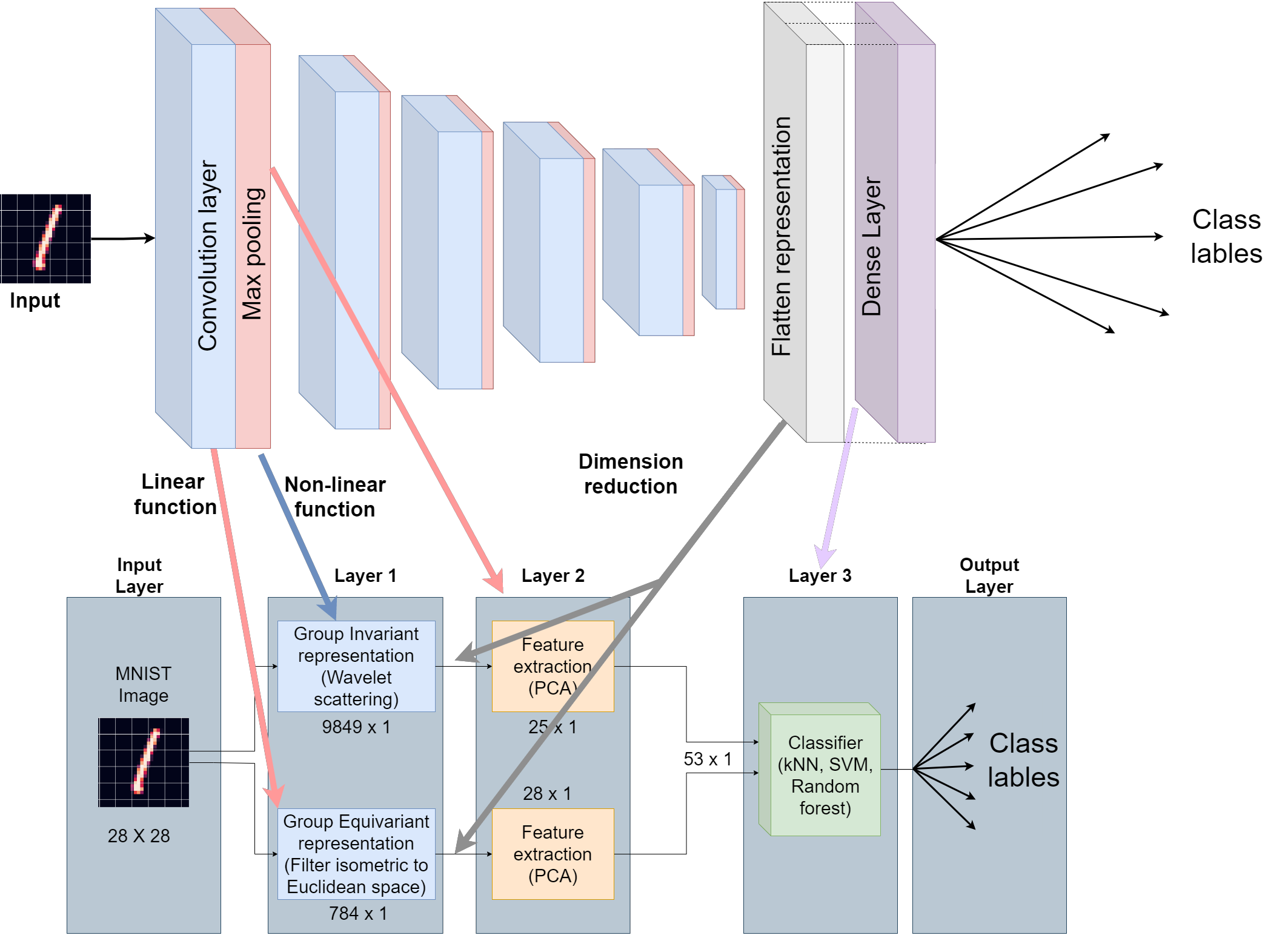}
  \caption{A prototype for representational interpretable CNN.}
  \label{fig:cnn-model}
  \label{fig:model}
\end{figure*}

Fig.~\ref{fig:cnn-model} shows a general architecture of a conventional CNN. It also depicts how the three layers of the proposed interpretable CNN correlate with the different layers in a conventional CNN. The convolution layer is represented in Layer 1 of the interpretable CNN. The non-linearity property of max pooling layers is performed by the modulus at the end of the invariant transformation. The dimension reduction task of max-pooling layers is performed by Layer 2. Flattening of the data is performed between Layer 1 and Layer 2. The classification performed by the dense layer is replaced with any interpretable classifier like kNN, SVM, or RFs. This way we create an interpretable CNN that implements similar operations to conventional CNN.

\subsubsection{Layer 1: Representation}
\label{Sec:GIFE}
As shown in Fig.~\ref{fig:model}, the first is the image representation layer. The input to this layer is an image, and the invariant and equivariant representations are the outputs. Wavelets and group equivariant operators are implemented to obtain invariant and equivariant representations, respectively. Note that invariance representation is ideal for increasing generalization ability at the same time, equivariant representation helps control the extent of transformation influence and how one class differentiates from another. This is a key characteristic needed to address multi-class classification for large scale problems.

Invariance to a transformation means the representation of an image and its transformed image are equal. Fig.~\ref{fig:space}, $\phi(x)$ be a representation, i.e., \{$\phi(x): X \to \mathbb{R}^{n \times n}\}|\{x \in \mathbb{R}^{n \times n}\} \in X$ is a set of all images. Consider $G$ be the group of global symmetries acting on the $X$. Now, $\phi(x)$ is invariant to the action of $g \in G$, i.e.,

\begin{equation}
  g.x=x_g
  \end{equation}
\begin{equation}
  \phi(x)=\phi(x_g),
  \end{equation}where, $x_g$ is transformed $x$ with the action of $g$. Let $\Phi_x$ be an invariant representation of $x$ under group $g$ be defined as,

\begin{equation}
  \Phi_x = {\Phi_x}_g.
  \label{eq:group}
\end{equation}

See that the group invariant non-linear representation equation (\ref{eq:group}) captures the common features among all the data points belonging to a particular class. In this work, we obtain an invariant representation of the data by applying scattering wavelet transform (SWT). It exploits the fact that the wavelet transformations are invariant to small diffeomorphisms. However, they are covariant to translations, and to obtain a translation invariant, a nonlinear operation is performed at the end of each layer like modulus. The initial layer contained in SWT is a scaling operator resulting in the output $x.\phi$. The $n^{th}$ order coefficients are $||x*\psi_1|*\psi_2|...|*\psi_n|.\phi$. We conducted experiments to determine the scattering order at which optimal representation is obtained. We found that at scattering order $1$, the representation has the highest invariance. So, the best representation is obtained when no cascade of wavelets is applied. The scattering transform has an output at the end of each layer called $n^{th}$ order coefficients for $n^{th}$ layer.  The coefficients are used as invariant representations in the proposed model (Fig:~\ref{fig:model} ). We employ Morlet wavelet since it is considered closest to human audio and visual perception.

\label{sec:equ}
Now referring to Fig.~\ref{fig:model}, the equivariant representation is now obtained in the Layer $1$ in order to regulate the generalization in the features extracted to represent the image. Equivariant features contain information about the underlying symmetries in the data. Let an image $x$ be represented by the function $x(i,j)$, and the equivariant representation after applying the operator is $\phi(i,j)$. A group equivariant operator acts as a function between the two spaces $(x,\phi)$ that act by the same symmetry groups. Here, We propose to preprocess the data to obtain the equivariant representation that is given as input to a classifier. We implement isometry equivariant non-expansive operators (IENEO), that are convolution operators parametric with respect to $P$, mapping continuous function with compact support $x:\mathbb{R}^2 \to \mathbb{R}$ to a continuous compactly supported function $\phi:\mathbb{R}^2 \to \mathbb{R}$.

\begin{figure}[]
\centering
\includegraphics[scale=0.29]{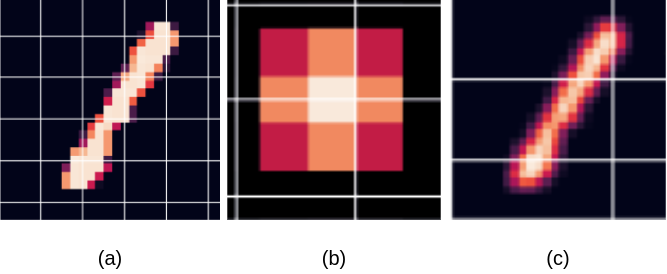}
\caption{Group equivariant (GE) representation: (a) an image, (b) a GE operator, and (c) GE representation of (a).}
\label{fig:visual}
\end{figure}

\begin{figure}[]
\centering
\includegraphics[scale=0.47]{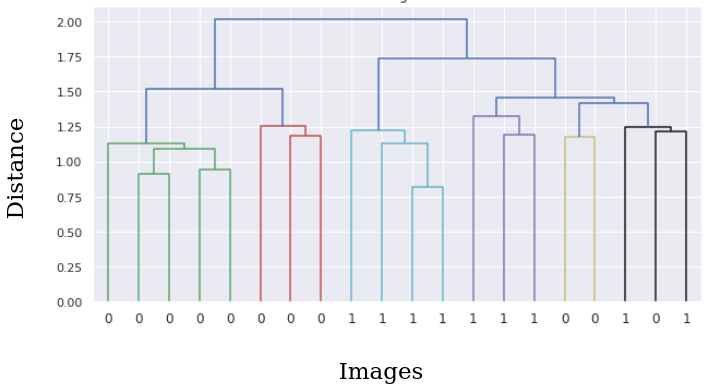}
\caption{Dendrogram of images from fashion MNIST.}
\label{fig:dend1}
\end{figure}

\begin{figure}[]
\centering
\includegraphics[scale=0.47]{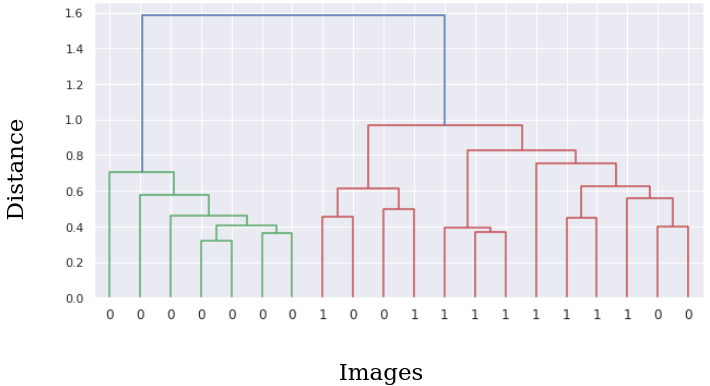}
\caption{Effect of GE representation on Fig.~\ref{fig:dend1}.}
\label{fig:dend2}
\end{figure}
\subsubsection{Layer 2: Dimension reduction }
\label{sec:pca}
Once the invariant and equivariant representations are obtained in Layer $1$, we perform principal component analysis (PCA) to further select the respective invariant and equivariant features in order to generate a compact yet complete representation. The PCA selects features by calculating eigenvectors of the covariance matrix and the higher eigenvalues are considered features. The eigenvalues are considered features extracted from the image representations. The eigenvectors of the training database are taken as a basis and the test data is represented using them. Once the eigenvectors are calculated based on the training data, we create an instance of the model with a user defined percentage of variance retained. We found that $80\%$ variance retention has the best results.

\subsubsection{Layer 3: Classifier}
Now that we have a set of compact features extracted from respective invariant and equivariant representations, we need to implement a learning model to classify images based on the compact features obtained after the dimension reduction. It is clear that these features encompass both the class information obtained through invariant representation and the information about the underlying symmetries of the data obtained through equivariant representation. Hence, they are now better equipped to classify the data (image) without generating various representations within a CNN as generally done in conventional CNNs. 
\subsubsection{Multi-class interpretable CNN}
In the proposed method, the wavelets and IENEOs can effectively generate group equivariant and invariant representations between two classes. Further studying of conventional multi-class CNNs would help us in creating multi-class group equivariant filters. That would extract specific inter class and intra class features across multiple classes. This would help us in designing interpretable CNN models for large scale vision problems.

\subsection{Representational insights from the implementation invariances}
As illustrated in Fig.~\ref{fig:cnn-model}, when an image of size $28 \times 28$ from the MNIST database is given to the prototype interpretable CNN (Fig.~\ref{fig:model}), a group invariant representation of the given image of size $9849 \times 1$ is obtained through SWT. Similarly, a group equivariant representation of size $784 \times 1$ is obtained by using IENEOs. Subsequently, features are extracted from the two representations using PCA resulting in $25$ invariant features and $28$ equivariant features (Fig.~\ref{fig:cnn-model}). These are finally given to the classifier.

We implement to obtain an equivariant representation of the data in the Euclidean space. Such representations are stable to symmetric transformations and are closer to images of a similar class. We select the operators that preserve the class information while decreasing variations in the data. To this end, we resort to a topological data analysis tool called persistent homology \cite{per}. This can be used to measure the topological features and shapes of a function. This is used to compare the performance of randomly generated operators. The effect of their transformations on the data is studied and operators that increase the inter-class distance and decrease the intra-class distance are selected. Fig.~\ref{fig:visual} illustrates the shape and effect of an operator selected while classifying the MNIST dataset. While selecting the equivariant operator, we introduce an extra step that builds a classifier using $70\% $ of training data and validated the model with the rest of the training data. The operator with the best performance is selected to obtain equivariant representation. This leads to the selection of an equivariant operator best suitable for a classifier. The impact of equivariant isometric representation can be observed in dendrograms shown in Fig.~\ref{fig:dend1} and Fig.~\ref{fig:dend2}. One can observe from the dendrograms that there are differences in the distances among the clusters after obtaining the group equivariant representations. Similar images cluster far earlier (Fig.~\ref{fig:dend2}) than in the case of unprocessed images (Fig.~\ref{fig:dend1}). Further, the number of clusters drastically decreases after obtaining an equivariant representation of the data (Fig.~\ref{fig:dend2}). 
    
In our experiments, we implement kNN, SVM, and random forest to classify the data. Through invariant representation, we obtain common features of a class which helps us identify the class label for unknown data. On the other hand, through a non-expansive equivariant operator, we obtain features that are stable to transformations in the data. It is interesting to see that the features extracted from these two representations as an input help improve the accuracy of conventional classifiers, which do not have an abundant internal representation of the data, unlike neural network classifiers. The proposed method helps to address the representational aspects of the data. In particular, when SVM is implemented within the framework Fig.~\ref{fig:cnn-model} its performance is at par with AlexNet (see section Sec.\ref{Sec:results}). While the representation part of the framework imitates the convolution layers in a CNN, the SVM within the network Fig.~\ref{fig:cnn-model} imitates the fully connected layers and softmax of the CNN.

\subsubsection{A measure of Equivariance map}
\label{sec:measure}
The internal representation of the data plays a crucial role in a CNN performance. Now, we perform auditing to justify the proposed framework. Since extreme transformations result in unrecognizable images or changes in class labels. In the proposed method, invariant and equivariant representations are obtained in Layer 1. Dimension reduction is performed on this representation to obtain features. These features can be evaluated by measuring the equivariance capacity. We estimate the equivariant map and evaluate the accuracy at which the impact of transformations is predicted by the map. 

\begin{figure}
\centering
\includegraphics[scale=0.4]{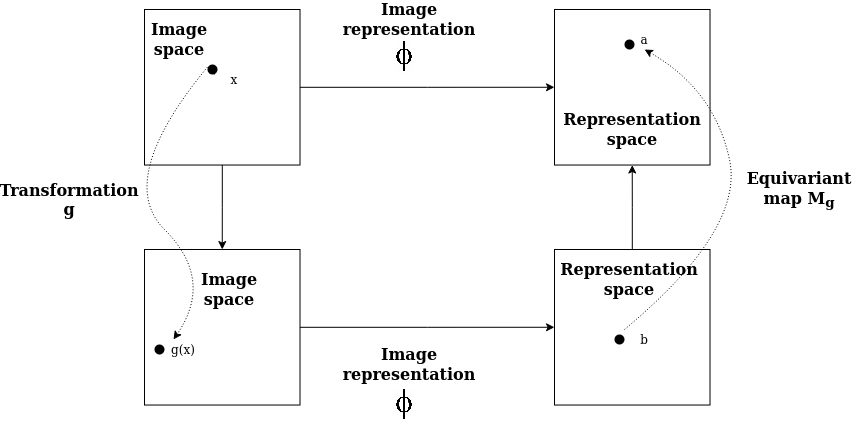}
\caption{An illustration of representation equivariance.}
\label{fig:equi}
\end{figure}

An image representation $\phi(x)$ is considered equivariant with respect to a transformation $g$ if a map $M_g$ exists such that,

\begin{equation}
  \phi(g(x)) = M_g(\phi(x)) \textnormal{   } \forall x \in X.
  \label{eq:mg}
\end{equation}
\noindent This is illustrated in the Fig.~\ref{fig:equi}. If the representation is invariant to a particular transformation then the equivariant map $M_g$ will be identified. The existence of inverse of $\phi$ and closure with respect to transformations group $G$ are sufficient conditions for the existence of the equivariant map $M_g$ \cite{measure,valdi}.

Equivariance of a representation to a transformation can be proved by calculating the equivariance map. In our work, (Fig:~\ref{fig:model}), $M_g$ is calculated through linear regression. The approximation of the equivariant map is calculated through

\begin{equation}
  \widehat{M_g}= \phi(g(x)){\phi(x)}^{-1},
  \label{eq:reg}
\end{equation}
\noindent where, $\widehat{M}_g$ is the approximation of $M_g$ calculated through the regularized linear regression. After calculating $\widehat{M}_g$, we evaluate the equivariance of the representation as follows,
\begin{equation}
 \textnormal{if } \phi(g(x) = \widehat{M_g}(\phi(x)) \textnormal{ then, } \\
  \phi(x)=\widehat{M_g}(\phi(g^{-1}(x))) .
\end{equation}

\begin{equation}
  E_q = \frac{1}{n} \sum_{\forall x} |\phi(x)-\widehat{M_g}(\phi(g^{-1}(x))) |.
  \label{eq:cal}
\end{equation}
\noindent Here $E_q$ indicates the equivariance capability of the function $\phi$. The lower the value of $E_q$, the higher the equivariant capacity of a representation.

\section{Simulation}
\label{Sec:results}
This section demonstrates simulation results on various databases, using the prototype representational interpretable CNN and performance is compared with AlexNet along with conventional baseline classifiers including kNN, RF, and SVM. 

One can observe from Table~\ref{tab:quant2}, that when the available data is small (plant pathology and cancer detection), our approach has higher accuracy. Table~\ref{tab:quant} details the improvement in the classification by implementing the proposed representation in kNN, SVM, and random forest classifiers within the interpretable CNN. We can observe that the SVM with RBF kernel has a better improvement. The proposed representation not only improves the performance of conventional classifiers but also decreases the computations as the dimension is reduced through PCA. We interestingly notice that when the variance in the data is high, then the extracted features as input have significantly improved the accuracy of a classifier. This can be observed by comparing CIFAR-10 results (Table:~\ref{tab:quant}).

\begin{table}[h]
\caption{Classification accuracy of proposed prototype (Fig:~\ref{fig:model}, and the AlexNet (CNN) \cite{alex}.}
\label{tab:quant2}
\begin{center}
\begin{tabular}{|l||c||c||c|}
\hline
Datasets                                                    & \begin{tabular}[c]{@{}c@{}}Proposed\\ model \\ (Fig:~\ref{fig:model}) \end{tabular} & {\begin{tabular}[c]{@{}c@{}}AlexNet\\ (CNN)\\ \cite{alex}\end{tabular}} \\ \hline \hline
\begin{tabular}[c]{@{}l@{}} MNIST                 \cite{mnist}     \end{tabular}                                  & \textbf{0.99}                                                                 & \textbf{0.99}                                                                         \\ \hline
\begin{tabular}[c]{@{}l@{}} FMNIST            \cite{fmnist}      \end{tabular}                                     &       \textbf{0.98}                                                                                             & \textbf{0.98}                                                                         \\ \hline
\begin{tabular}[c]{@{}l@{}} CIFAR-10          \cite{cifar}      \end{tabular}                                     & 0.93                                                                                                                                                 & \textbf{0.94}                                                                         \\ \hline
\begin{tabular}[c]{@{}l@{}}Skin lesions  \cite{skin} \end{tabular}      & \textbf{0.68}                                                                                                                                          & 0.65                                                                         \\ \hline
\begin{tabular}[c]{@{}l@{}}Plant pathology \cite{plant}\end{tabular}  & \textbf{0.73}                                                                                         & 0.55                                                                         \\ \hline
\begin{tabular}[c]{@{}l@{}}Cats vs Dogs \cite{dogs_cats} \end{tabular}     & 0.71                                                        & \textbf{0.81}                                                                         \\ \hline
\begin{tabular}[c]{@{}l@{}}Cancer \\detection \cite{cancer1} \end{tabular} & \textbf{0.71} & 0.55                                                                         \\ \hline
\end{tabular}
\end{center}
\end{table}

\begin{figure}[h]
  \centering
  \includegraphics[scale=0.8]{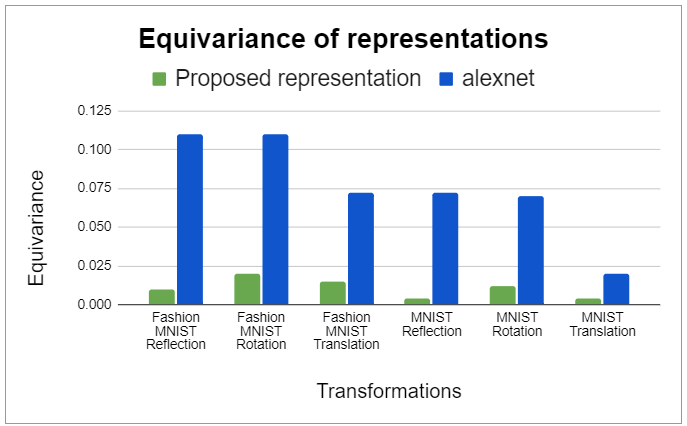}
\caption{ Equivariance of proposed representation (Layer 2 of Fig:~\ref{fig:model} ) Vs AlexNet (CNN).}
  \label{fig:equivariance}
\end{figure}

Following this, we implement the algorithm to measure the equivariance of the proposed representation (Layer 2 output in Fig:~\ref{fig:model}) by calculating the equivariant map \cite{measure} using the equation (\ref{eq:cal}). For comparison, we measure the equivariance of conventional CNN's internal representation. The output of the last layer before the softmax dense layer of AlexNet is considered as its internal representation. The impact of a transformation on the representation can be observed in Fig:~\ref{fig:equivariance}. It depicts the equivariance capability of our proposed representation vs AlexNet internal representation. The $E_q$ values of each representation with respect to group actions i.e., translation, rotation, and reflection are calculated. The lower the value of $E_q$, the higher the equivariance capability. One can observe that our proposed representation is equivariant to symmetric transformations like translation, rotation, and reflection. This contributes to the generalization ability of interpretable CNN. For translation, the average of all the equivariance measures in both directions is considered. We have only calculated for $0$ to $5$ pixel translation as the higher translations would result in the image is unrecognizable. For rotation, the average of clockwise rotations from $10$ to $90$ degrees is calculated.

\begin{table*}[h]
\caption{Accuracy analysis.}
\label{tab:quant}
\begin{center}
\begin{tabular}{|l||c|c||c|c||c|c|}
\hline
Datasets                                                    & \begin{tabular}[c]{@{}c@{}}Baseline\\ kNN\end{tabular} & \begin{tabular}[c]{@{}c@{}}kNN  within\\unboxed\\CNN model\end{tabular} & \begin{tabular}[c]{@{}c@{}}Baseline\\ RF\end{tabular} & \begin{tabular}[c]{@{}c@{}}RF within\\unboxed\\CNN model\end{tabular} & \begin{tabular}[c]{@{}c@{}}Baseline\\ SVM\end{tabular} & \begin{tabular}[c]{@{}c@{}}SVM within\\unboxed\\CNN model\end{tabular} \\ \hline \hline
\begin{tabular}[c]{@{}l@{}}MNIST \cite{mnist} \end{tabular}                                             & 0.95                                                    & \textbf{0.99}                                                                 & \textbf{0.99}                                                   & \textbf{0.99}                                                                & \textbf{0.99}                                                   & \textbf{0.99}                                                                 \\ \hline
\begin{tabular}[c]{@{}l@{}}FMNIST \cite{fmnist} \end{tabular}  & 0.94                                                    & \textbf{0.98}                                                                 & 0.96                                                   & \textbf{0.98}                                                                & 0.96                                                    & \textbf{0.98}                                                                 \\ \hline
\begin{tabular}[c]{@{}l@{}}CIFAR-10  \cite{cifar} \end{tabular}                                                    & 0.50                                                    & \textbf{0.88}                                                                 & 0.74                                                   & \textbf{0.86}                                                                & 0.68                                                    & \textbf{0.93}                                                                 \\ \hline
\begin{tabular}[c]{@{}l@{}}Skin lesions \cite{skin}\end{tabular}      & 0.63                                                    & \textbf{0.64}                                                                 & 0.63                                                   & \textbf{0.65}                                                                & 0.66                                                    & \textbf{0.68}                                                                 \\ \hline
\begin{tabular}[c]{@{}l@{}}Plant  pathology \cite{plant}\end{tabular}  & 0.50                                                    & \textbf{0.56}                                                                 & 0.55                                                   & \textbf{0.65}                                                                & 0.57                                                    & \textbf{0.73}                                                                 \\ \hline
\begin{tabular}[c]{@{}l@{}}Cats vs Dogs \cite{dogs_cats}\end{tabular}     & 0.59                                                    & \textbf{0.62}                                                                 & 0.60                                                   & \textbf{0.64}                                                                & 0.62                                                    & \textbf{0.71}                                                                 \\ \hline
\begin{tabular}[c]{@{}l@{}}Cancer detection \cite{cancer1}\end{tabular} & 0.62                                                    & \textbf{0.67}                                                                 & 0.66                                                   & \textbf{0.67}                                                                & 0.65                                                    & \textbf{0.71}                                                                 \\ \hline
\end{tabular}
\end{center}
\end{table*}

\section{Concluding Remarks and lessons learnt}

In this work, we have performed a post-hoc analysis of CNN model with the aim to create an interpretable prototype to better understand the representational aspects of CNN. In this regard, we have studied the structures and properties of conventional CNNs and found that the generalization ability stems from its equivariant internal representation. Based on this, we suggest a representation method that emulates equivariance and invariance to group transformation properties. The equivariant representation is obtained by implementing filters isometric to Euclidean space and the invariant representation is obtained through wavelet scattering transform. Selected features (PCA) extracted from the equivariant and invariant representations are then employed for classification. It is further validated through simulations for binary classification that the proposed interpretable CNN performs better when training data is scarce and presents a transparent neural model.

Invariant representation captures the class information of the data while equivariant representation captures the symmetries of the data. With this, the proposed representation is better suited to improve the accuracy of a classifier for the data with a high variance within the same class. One can also observe that the introduction of both invariance and equivariance representation reduces the solution space to be explored while constructing a neural network based classifier. This way the method can be used for preprocessing large data for classification when the data is scarce. This is reflected in simulations when classifiers that are sensitive to the local structure of the data like SVM are implemented in Layer 3. Finally, we have measured the equivariance to various transformations and observed that the proposed representation is robust to symmetric transformations of the input image for two-class problems. Our work would help in building interpretable CNN models for multi-class large scale problems in the future.

Future research directions include working on, equivariant representations and creating a network of such representations. In the next phase of the research, we hope to study the effects of interpretable representations of state-of-the-art neural networks like ResNet-50. We also hope to extend this work for multi-class classification using the critical insights gained from improving inter-class and intra-class accuracies in this binary case. Given that the invariant representations and equivariant representations are independent, one may also come up with a method that parallelly computes representations of various images to improve the speed of the classification for real-time large scale industrial applications.

\bibliographystyle{unsrt}  
\bibliography{references}  

\appendix

\section{Research Methods}

All the experiments are conducted on an Intel core i7-9th Gen 16GB RAM, CPU@2.2GHz with Nvidia Geforce GTX 1660Ti 4GB graphics. The experiments are conducted with two classes in the dataset and each experiment is run three times. The average accuracy of the three runs is presented in the Sec:~\ref{Sec:results}. The accuracy is calculated using the confusion matrix as follows,
\begin{center}
        \begin{equation}
        Accuracy = \frac{(TN+TP)}{(TP+TN+FN+FP)},\label{eq:5}
    \end{equation}
\end{center}
where TN : True negatives, TP : True positives, FP : False negatives, FN : False positives. The k value for kNN is empirically set to as 11 since odd numbers are preferred for classification with two classes. We also observe that the performance of kNN is stable to change in k value. RBF kernel is used for implementing the SVM algorithm. Besides, we implement linear and polynomial kernels and we found that RBF exhibits better performance among the three. In the case of random forest, 100 trees are implemented. We discerned that an increase in the number of trees has not improved the performance.

\end{document}